\documentclass[runningheads, envcountsame, a4paper]{llncs}
\pdfoutput=1 
\usepackage{amsmath}
\usepackage{amssymb}
\usepackage[misc]{ifsym}
\usepackage{graphicx}
\usepackage[official]{eurosym}
\usepackage[hyphens]{url}
\usepackage[breaklinks=true,colorlinks]{hyperref}

\begin{document}
\title{Temporal Smoothing for 3D Human Pose Estimation and Localization for Occluded People}
\titlerunning{Temporal Smoothing for 3D Human Pose Estimation}
%
\author{M. V\'eges \and A. L\H{o}rincz}

\toctitle{Temporal Smoothing for 3D Human Pose Estimation and Localization for Occluded People}
\tocauthor{M.~V\'eges and A.~L\H{o}rincz}

\authorrunning{M. V\'eges \and A. L\H{o}rincz}
%
\institute{E\"otv\"os Lor\'and University, Budapest, Hungary \\
\email{$\{$vegesm,lorincz$\}$@inf.elte.hu}}
\maketitle              
\setcounter{footnote}{0}
\begin{abstract}
In multi-person pose estimation actors can be heavily occluded, even become fully invisible behind another person. While temporal methods can still predict a reasonable estimation for a temporarily disappeared pose using past and future frames, they exhibit large errors nevertheless. We present an energy minimization approach to generate smooth, valid trajectories in time, bridging gaps in visibility. We show that it is better than other interpolation based approaches and achieves state of the art results. In addition, we present the synthetic MuCo-Temp dataset, a temporal extension of the MuCo-3DHP dataset. Our code is made publicly available.\footnote{\url{https://github.com/vegesm/pose_refinement}}

\end{abstract}

\section{Introduction}
The task of 3D human pose estimation is to predict the coordinates of certain body joints based on an input image or video. The potential applications are numerous, including augmented reality, sport analytics and physiotherapy. In a multiperson settings, 3D poses may help analyzing the interactions between the actors.

Recent results on the popular Human3.6M database \cite{h36m} are starting to saturate \cite{kocabas2019epipolar,integralPose} and there is an increasing interest in more natural settings. The standard evaluation protocol in Human3.6M uses hip-relative coordinates. It makes the prediction task easier, as the localization of the pose is not required. However, in multi-person settings the distance between the subjects can be important too. This led to the introduction of absolute pose estimation \cite{moon2019camdistance,veges2019depthpose}, where the relative pose prediction is supplemented with the localization task.

Note that this problem is underdefined when only a single camera provides input. It is impossible to estimate the scale of the real scene based on one image, if semantic information is not available. Therefore, we use a scale-free error metric that calculates the prediction error up to a scalar multiplier. Additionally, our method is temporal, that lets us discern special cases such as jumping. An image-based method does not have enough information to decide, whether the jumping person is closer to the camera or higher above the ground.

\begin{figure*}[t]
\centering
\includegraphics[width=0.9\textwidth]{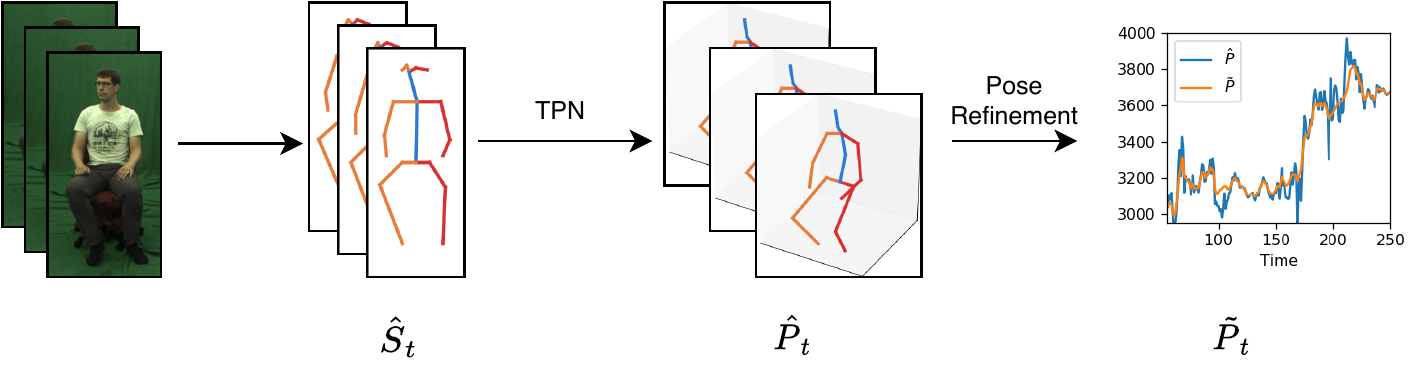}
\caption{\textbf{Overview of our algorithm.} First 2D pose estimations ($\hat{S}_t$) are generated for all frames. Then the Temporal PoseNet (\textit{TPN}) converts the 2D estimates to initial 3D poses ($\hat{P}_t$). Finally, the pose refinement step smoothes out the trajectories, producing the final 3D pose estimates ($\tilde{P}_t$).}
\label{fig:overview}
\end{figure*}

Temporal methods can exploit information in neighboring frames. When a person disappears for a short time (for instance someone walks in front of him/her), these methods could fill in the void based on the previous and following frames. Still, the output is often noisy and even a simple linear interpolation yields better results (see Table \ref{tbl:interpol}).

To overcome the occlusion problem, we propose two contributions. First, we introduce the MuCo-Temp synthetic dataset. It is a temporal extension of the MuCo-3DHP dataset \cite{mehta2018single_shot}. The latter database is commonly used as the training set for MuPoTS-3D \cite{mehta2018single_shot}. The MuPoTS-3D dataset is a multi-person 3D human pose database, containing 20 video sequences. It consists of a test set only so the authors of \cite{mehta2018single_shot} introduced MuCo-3DHP, that is generated from the single person MPI-INF-3DHP dataset \cite{mehta}. Each image in MuCo-3DHP is a composition of four poses from MPI-INF-3DHP. This new synthetic dataset contains occlusions typical to multi-person settings but has images only while our method is temporal. 
Thus, we created the MuCo-Temp dataset, containing videos composited from sequences in the MPI-INF-3DHP database. To keep compatibility, we used the same algorithm as the authors of MuCo-3DHP.

Our second contribution is an energy minimization based smoothing function, targeting specifically those frames where a person became temporarily invisible. It adaptively smoothes the prediction stronger at frames where the pose is occluded and weaker when the pose is visible. This way large noises during heavy occlusions can be filtered without  `over-smoothing' unoccluded frames. The method does not require additional training and can be applied after any temporal pose estimation algorithm.

To summarize, we introduce an approach to predict 3D human poses even when the person is temporarily invisible. It achieves state-of-the-art results on the MuPoTS-3D dataset, showing its efficiency.

\section{Related Work}
\textbf{3D Pose estimation} Before the widespread usage of deep learning, various approaches were explored for 3D human pose prediction, such as conditional random fields \cite{belagiannis2014pictorialpose} or dictionary based methods \cite{ramakrishna2012dictionary}. However, recent algorithms are all based on neural networks \cite{3dbaseline,pavlakos2018ordinal,integralPose}. The primary difficulty in 3D human pose estimation is the lack of accurate in the wild datasets. Accurate measurements require a studio setting, such as Human3.6M \cite{h36m} or MPI-INF-3DHP \cite{mehta}. To overcome this, several approaches were proposed to use auxiliary datasets. Zhou et al. \cite{zhou2017} uses 2D pose datasets with a reprojection loss. Pavlakos et al. \cite{pavlakos2018ordinal} employed depthwise ordering of joints as additional supervision signal. An adversarial loss added to the regular regression losses ensure the plausibility of generated 3D poses \cite{drover_2dto3d}. It requires no paired 2D-3D data.

More importantly, Martinez et al. \cite{3dbaseline} introduced a two step prediction approach: first the 2D pose is predicted from the image, then the 3D pose is estimated solely from the 2D joint coordinates. This places the task of handling image features on the 2D pose estimator, for which large, diverse datasets exist \cite{mscoco}. The 2D-to-3D regression part can be a simple feed-forward network \cite{3dbaseline}, one that uses recurrent layers \cite{fang2018posegrammar,lee2018pLSTM} or a graph convolutional network \cite{zhaoCVPR19semantic}. Additionally to employing off-the-shelf algorithms \cite{hrnet}, the joint training of the 2D pose estimator and the 2D-to-3D part is possible via soft-argmax \cite{Luvizon2018softargmax,integralPose}. The soft-argmax function is a differentiable approximation of the argmax operation that lets gradients flow to the 2D estimator. In our paper we follow the two step approach with a pretrained 2D pose estimator \cite{hrnet}.

\textbf{Temporal methods} While 3D poses can be inferred from an image only, for videos temporal methods provide better performance than simple frame-by-frame approaches. A natural idea is to use recurrent layers over per-frame estimates \cite{Hossain2017temporal}. However, even with LSTMs, RNN based methods exploit only a small temporal neighborhood of a frame. This can be solved with 1D convolution \cite{pavllo2019videopose3d}. To increase the receptive field without adding extra parameters, the authors used dilated convolutions. Additionally, bundle adjustment \cite{arnab2019meshbundle} was used to refine body mesh prediction in time. Our work is closest to \cite{arnab2019meshbundle}, however their method smoothes coordinates for visible poses only and does not handle occlusions.

Instead of working directly on joint coordinates, Lin et al. first decomposed the trajectories with discrete cosine transform \cite{lin2019trajectory}.  
In \cite{wang2020motionloss} the authors introduced a special motion loss that encourages realistic motion paths. Finally, Cai et al \cite{cai2019stcnn} use a spatio-temporal graph convolutional network. The standard uniform convolution was changed such that the kernels became different for different joints. 

\textbf{Localization and pose estimation} Most work focuses on relative pose estimation: the joint coordinates are predicted relative to a root joint, usually the hip. This is sufficient for many use cases, however for multi-person images, the locations of the actors might be important too. The joint estimation of location and relative pose is also called absolute pose estimation \cite{veges2019depthpose}. In \cite{veges2020wdspose}, the authors propose a direct regression approach that learns from RGB-D data as well. 
Moon et al. \cite{moon2019camdistance} uses two separate networks to predict the location and the 3D pose.

Additional performance can be gained by using multiple cameras. The authors of \cite{qiu2019crossview} predict the joint locations based on many viewpoints. They employ a modified pictorial structure model that recursively increases its resolution around the joint locations, improving accuracy while keeping speed. 

\textbf{Occlusions} Occlusions can be grouped in two categories: self-occlusions and occlusions caused by other objects or people. In typical single-pose estimation datasets, only the first type arises. Mehta et al. introduces Occlusion Robust Heatmaps, that store joint location coordinates as a 2D heatmap \cite{mehta2018single_shot}. Augmenting datasets with synthetic occlusions also helps. Sarandi et al. overlaid rectangles and objects from MS-COCO on individual frames \cite{sarandi2018eccv_winner}. Cylindrical man models can be used to predict visibility from 3D joint coordinates \cite{cheng2019occlusionaware}. The visibility flags combined with 2D joint detection score can form the basis of an effective regularizer.

\begin{figure*}[t]
\centering
\includegraphics[width=0.97\textwidth]{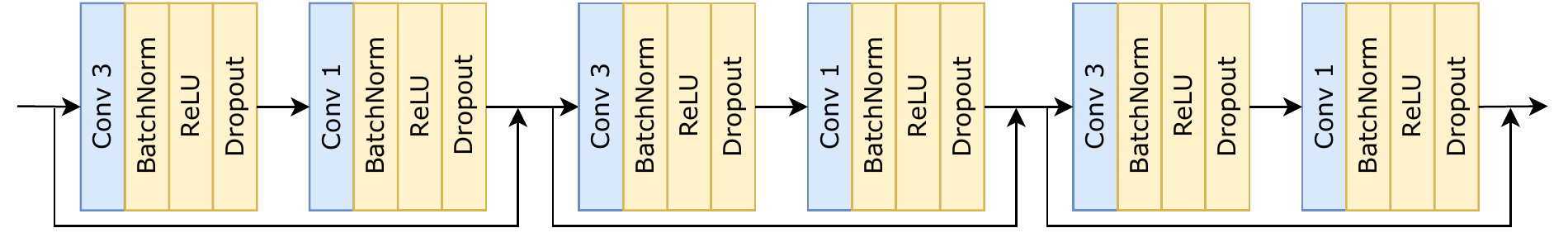}
\caption{\textbf{Temporal PoseNet architecture.} The network has three residual blocks, each block contains two convolutional layer followed by BatchNorm and Dropout layers. The activation function was ReLU.}
\label{fig:tpn}
\end{figure*}

\section{Method}
Our algorithm predicts the coordinates of the joints of a person in a camera relative coordinate system, where the origin is at the camera center and the axes are parallel to the camera plane. The estimation of coordinates is separated into two tasks: localization and relative pose estimation, similar to previous work \cite{veges2019depthpose,moon2019camdistance}. That is, we predict the location of the root joint (the hip) and the location of the other joints relative to the hip. We denote the former $P^{loc}$, the latter $P^{rel}$.

The general outline of our algorithm is as follows (see Figure \ref{fig:overview}): first $\hat{S}_{k,t}$, the 2D position of the $k$th person on frame $t$, is predicted using an off-the-shelf algorithm \cite{hrnet}. Then, a temporal pose network (\textit{TPN}) predicts the 3D location and pose of the person ($\hat{P}^{loc}_{k,t}$ and $\hat{P}^{rel}_{k,t}$ respectively), see Section \ref{sec:tpn} for details. Finally, the pose refinement step smoothes the predictions of TPN, producing the final predictions $\tilde{P}^{loc}_{k,t}$ and $\tilde{P}^{rel}_{k,t}$ (see Section \ref{sec:refiner}). The TPN and the pose refinement are run for each person separately. For brevity, we drop the index $k$ in the discussions below.

\subsection{Temporal PoseNet} \label{sec:tpn}
The Temporal PoseNet takes the 2D joint coordinates produced by the 2D pose estimator and predicts the 3D location and pose. To be robust against the different cameras in the training and test set, the 2D joints are normalized by the inverse of the camera calibration matrix. Then, for frame $t$ the joint locations are predicted based on a window of size $2w+1$:
$$\left[\hat{P}^{loc}_{t}; \hat{P}^{rel}_{t}\right] = f_{TPN}\left([K^{-1}\hat{S}_i]_{i=t-w}^{t+w}\right),$$
where $K$ is the intrinsic camera calibration matrix and $f_{TPN}$ is the Temporal PoseNet. 

The network architecture of TPN was inspired by \cite{pavllo2019videopose3d}, Figure \ref{fig:tpn} shows an overview. It is a 1D convolutional network, that has three residual modules. Each module contains a convolution of size 3 followed by a convolution of size 1. The convolutional layers are followed by a Batch Normalization layer, a ReLU activation function and a Dropout layer. 
The input and the output of the network are normalized by removing the mean and dividing by the standard deviation.

The loss function is the $\ell_1$ loss, to be robust against outliers. The final loss is:
\[
L_{TPN}=\sum_t\left\Vert\hat{P}^{loc}_{t}-P^{loc}_{t}\right\Vert_1 + \left\Vert\hat{P}^{rel}_{t}-P^{rel}_{t} \right\Vert_1,\]
where $P^{loc}_{t}$ and $P^{rel}_{t}$ are the ground-truth location and the relative pose coordinates for frame $t$.

\subsection{Pose Refinement} \label{sec:refiner}
While the input of the TPN is a large temporal window of frames, it still makes mistakes, especially during heavy occlusions. We propose an energy optimization based smoothing method that adaptively filters the keypoint trajectories. The objective function is calculated separately for the location and relative pose. It is formulated as:
\begin{equation}\label{eq:obj}
E_{ref}(\tilde{P})=vE_{pred}(\tilde{P})+(1-v)\lambda_1E_{smooth}(\tilde{P},\tau_1)+\lambda_2E_{smooth}(\tilde{P},\tau_2),
\end{equation}
where $\tilde{P}$ is is either $\tilde{P}^{loc}$ or $\tilde{P}^{rel}$, indicating whether the objective function optimizes the location or the relative pose; $v$ is the visibility score, $E_{pred}$ is an error function ensuring the smoothed pose is close to the original prediction, and $E_{smooth}$ is a smoothness error. 

The parameter $\tau$ of $E_{smooth}$ controls the smoothing frequency. A small value corresponds to a low-pass filter with a high frequency threshold, while a large value corresponds to a lower frequency threshold (see below the definition of $E_{smooth}$). We choose $\tau_1\gg\tau_2$.

The visibility score $v$ is close to 1 when the pose is visible, and to 0 when the pose is occluded. If the person is fully invisible, $v$ is set to 0. Thus, in Equation~\eqref{eq:obj}, when the pose is visible, only $E_{pred}$ is taken into account and the (stronger, $\tau_1$ parametrized) smoothing is ignored. In other words, when the pose was visible, the objective function does not override the predictions of TPN. On the other hand, when the pose is heavily occluded, only $E_{smooth}$ is active and the optimization ensures that the predicted pose will be smooth.

The prediction error is
\[E_{pred}\left(\tilde{P}\right)=\sum_t\min\left(\left\Vert\tilde{P}_{t}-\hat{P}_{t}\right\Vert_2^2,m\right),\]
where $\hat{P}_t$ is either the location or the pose predicted by the Temporal PoseNet at frame $t$, and $\tilde{P}_t$ is the target of the optimization. The $\min$ function ensures that large outliers do not affect the objective function. This is essential because the predicted pose $\hat{P}$ can be noisy when the person is occluded. $m$ is a fixed parameter.

The smoothing error is a zero velocity loss \cite{arnab2019meshbundle}, and $\tau$ controls the scale of the smoothing:
\[E_{smooth}\left(\tilde{P},\tau\right)=\sum_t\left\Vert\tilde{P}_{t}-\tilde{P}_{t-\tau}\right\Vert_2^2.\]
The error encourages that the pose changes smoothly over time. If $\tau$ is small, then the filtering works in a small window, removing local noise. If $\tau$ is large, the error function looks at a larger timescale.

Finally, the visibility score $v$ is the confidence score predicted by the 2D pose estimator. Since the 2D estimator returns per-joint confidences, these are averaged. Finally, a median filter is applied on $v$.

The total energy function evaluates Equation \eqref{eq:obj} both for the location and relative position:
\[E_{total}(\tilde{P}^{loc})=E_{ref}(\tilde{P}^{loc})+\lambda_{rel} E_{ref}(\tilde{P}^{rel})\]
where $\lambda_{rel}$ is a weighting parameter.

\section{Experiments}
\subsection{Datasets}
We evaluated our method on the multi-person 3D pose dataset MuPoTS-3D \cite{mehta2018single_shot}. It contains videos taken in indoor and outdoor environments. Since MuPoTS-3D contains only test sequences and has no training set, commonly the MPI-INF-3DHP \cite{mehta} and MuCo-3DHP \cite{mehta2018single_shot} databases serve as the training set. However, MuCo-3DHP contains single frames only, so temporal models can not be trained on it. Therefore, we created the MuCo-Temp synthetic dataset.

\subsubsection{MuCo-Temp Dataset}
The MuCo-3DHP database contains synthetic images, composited from frames of  MPI-INF-3DHP. The latter dataset was recorded in a green-screen studio, so segmenting of the actors is easy. Each synthesized frame contains four persons copied on a single image. Optionally, the background can be augmented with arbitrary images.

MuCo-Temp is using the same generation algorithm as MuCo-3DHP but it consists of videos instead of frames. Each video contains 4 person and 2000 frames. In total, we generated  77 videos for training. The validation set was created using the same process. We did not augmented the background, as the 2D pose estimator was already trained on a visually diverse dataset.

Our method was trained on the concatenation of the MPI-INF-3DHP and MuCo-Temp datasets. 

\subsection{Metrics}
We calculate various metrics to evaluate the performance of our algorithm. The following list briefly summarizes each:
\begin{itemize}
  \item \textbf{MRPE} or Mean Root Position Error, the average error of the root joint (the hip) \cite{moon2019camdistance}. 
  \item \textbf{MPJPE} or Mean Per Joint Position Error, the mean Euclidean error averaged over all joints and  all poses, calculated on \textit{relative} poses \cite{3dbaseline}.
  \item \textbf{3D-PCK} or Percentage of Correct Keypoints, the percentage of joints that are closer than 150mm to the ground truth. This metric is also calculated on relative poses only.
\end{itemize}
Since our method uses a single a camera, predicting the poses and locations of people is an underdefined problem. The above metrics do not take into account this, and calculate every error in mm. Therefore, we also include unit-less variants of the above metrics called \textit{N-MRPE} and \textit{N-MPJPE} \cite{rhodinSkiing}. The definition of N\nobreakdash-MRPE is 
\[\min_{s\in\mathbb{R}}\frac{1}{N}\sum_{k,t}\left\Vert s\tilde{P}^{hip}_{k,t}-P^{hip}_{k,t}\right\Vert^2_2,\]
where $\tilde{P}^{hip}_{k,t}$ and $P^{hip}_{k,t}$ are the predicted and ground-truth locations of the hip for person $k$ on  frame $t$. The total number of poses is $N$. The formula above finds an optimal scaling parameter $s$ to minimize the error. In other words, the scaling of the prediction comes from the ground truth. The scaling constant $s$ is calculated over all people and frames of the video: that is, the size of two predicted skeletons must be correct relative to each other, while the absolute size is still unknown.

MuPoTS has two kinds of pose annotations: universal and normal. The normal coordinates are the joint locations in millimeters. The universal coordinates are like the normal ones, but each person is rescaled from their hip to have a normalized height. Following previous work \cite{moon2019camdistance,veges2020wdspose}, the 3D\nobreakdash-PCK metric is calculated on the universal coordinates of MuPoTS\nobreakdash-3D while the (N-)MRPE and (N-)MPJPE errors are calculated on the normal coordinates. All the metrics are calculated over each sequence separately, then averaged.

\subsection{Implementation Details} \label{sec:imp-details}
The half-window size $w$ in the Temporal PoseNet was 40, thus the full input window length was 81 frames. The dropout rate was set to 0.25. The training algorithm was Adam with a learning rate of $10^{-3}$, decayed by a multiplier of 0.95 on each epoch. The network was trained for 80 epochs.

To avoid overfitting, the training set was augmented by scaling the 2D skeletons. When the focal length of the camera remains unchanged, this roughly corresponds to zooming the image. This step is essential, otherwise the TPN may overfit to the $y$ location of joints.

The optimization algorithm minimizing $E_{total}$ was Adam, with a learning rate of $10^{-2}$. The optimization ran for 500 iterations. The temporal timestep $\tau_1$ was 20, $\tau_2$ was 1, and the threshold $m$ was 1. The values of the weighting parameters: $\lambda_{rel}=0.1$, $\lambda_{1}=0.1$ and $\lambda_{2}=1$.

\setlength{\tabcolsep}{6pt}

\section{Results}
First, we compare the results of our model to the state of the art (Table~\ref{tbl:mupots-sota}). We improve on all metrics, however, this is not a fair comparison, since these models are single-frame algorithms, while ours is a temporal model. Also, MRPE and MPJPE errors can be calculated on detected poses only for these models

\begin{table}[h]
\centering
\caption{\textbf{Comparison with state-of-the-art} MPJPE and MRPE errors are in mm. *~Non-temporal methods. $\dagger$~Error is calculated on detected frames only. }\label{tbl:mupots-sota}
\begin{tabular}{lccc}
 & MRPE  & MPJPE &  3D-PCK \\
\hline
LCR-Net++ \cite{rogez2019lcrpp}* & - & -  &  70.6 \\
Moon et al. \cite{moon2019camdistance}* & $277^\dagger$ & -  & 81.8 \\ 

Veges et al. \cite{veges2020wdspose}* & $238^\dagger$ & $120^\dagger$  & 78.2 \\
\hline
\textbf{Ours}$^\dagger$ & $225^\dagger$ & $100^\dagger$  & $86.8^\dagger$ \\
\textbf{Ours} & \textbf{252} & \textbf{103}  & \textbf{85.3} \\
\hline
\end{tabular}
\end{table}

To have a fair comparison, we also test against other baselines (Table~\ref{tbl:interpol}). First, we use a simple linear interpolation for frames when a person is occluded~(\textit{Interpolation}). While on relative pose estimation metrics it achieves similar performance as our model, the localization performance is considerably worse (272 vs 252 in MRPE, a 7.4\% relative drop).

\begin{table}[h]
\centering
\caption{\textbf{Comparison with baselines} \textit{Interpolation} uses simple linear interpolation for unseen poses. \textit{1-Euro} applies a 1-Euro filter on interpolated poses. (N-)MPJPE and MRPE errors are in mm.}\label{tbl:interpol}
\begin{tabular}{lccccc}
 & MRPE  & MPJPE &  3D-PCK & N-MRPE & N-MPJPE \\
\hline
Ours w/o refine & 340 & 107  & 83.4  &307 & 113\\ 
Interpolation & 272 & 103 & 85,1 &  243 & 109 \\ 
1-Euro & 273 & 104 & 85.1 & 243 & 109 \\
\textbf{Ours} & \textbf{252} & \textbf{103}  & \textbf{85.3} & \textbf{221} & \textbf{108} \\
\hline
\end{tabular}
\end{table}

Additionally, we apply the 1-Euro filter \cite{casiez2012one_euro} on the interpolated poses (\textit{1-Euro}). This filter was applied in previous work to reduce high-frequency noises \cite{vnect,zou2020footskating}. It is noticeable, that the filter had no additional effect over the interpolation, the metrics either did not change or marginally worsened. That is, our smoothing procedure performs better than the 1-Euro method.

\subsection{Ablation Studies}

We performed an ablation study to confirm that each component of our algorithm contributes positively, the results are presented in Table~\ref{tbl:ablation}a. The \textit{Baseline} corresponds to the Temporal PoseNet only, trained on MPI-INF-3DHP. Adding MuCo-Temp improves on all performance metrics. Moreover, adding Pose Refinement further decreases errors by a large amount (MRPE goes from 340 to 252, a 25\% drop).

\begin{table}[h]
\centering
\caption{\textbf{Results of ablation studies.} a) Results when components are turned on sequentially. b) Errors calculated on visible poses only.}\label{tbl:ablation}
\begin{tabular}{lccccc}
\multicolumn{6}{l}{\textbf{a) Performance of components}} \\
\hline
 & MRPE & MPJPE &  3D-PCK & N-MRPE & N-MPJPE \\
\hline
Baseline & 372 & 116 & 81,2 & 332 & 122 \\
+ MuCo-Temp & 340 & 107  & 83.4 & 307 & 113 \\
+ Pose Refinement & 252 & 103  & 85.3 & 221 & 108 \\ 
\hline
\\
\multicolumn{6}{l}{\textbf{b) Results on visible poses}} \\
\hline
 & MRPE  & MPJPE &  3D-PCK & N-MRPE & N-MPJPE \\
\hline
Ours w/o refine & 227 & 100  & 86.5 & 191 & 106 \\ 
Ours & 225 & 100  & 86.6 & 189 & 106 \\ 
\hline
\end{tabular}
\end{table}

We hypothesized that our model improves the estimation of occluded poses. To show that the pose refinement process does not hurt the accuracy of visible joints, we evaluated the model on those poses, where the visibility score reached a threshold (0.1 in our experiments). The results are shown in Table~\ref{tbl:ablation}b. The addition of the refinement step changed the performance only marginally, indicating that a) our method improves prediction on heavily occluded poses b) does not decrease performance on unoccluded poses.

\subsection{Vertical Location and Depth}
\begin{figure*}[t]
\begin{center}
\includegraphics[width=0.8\textwidth]{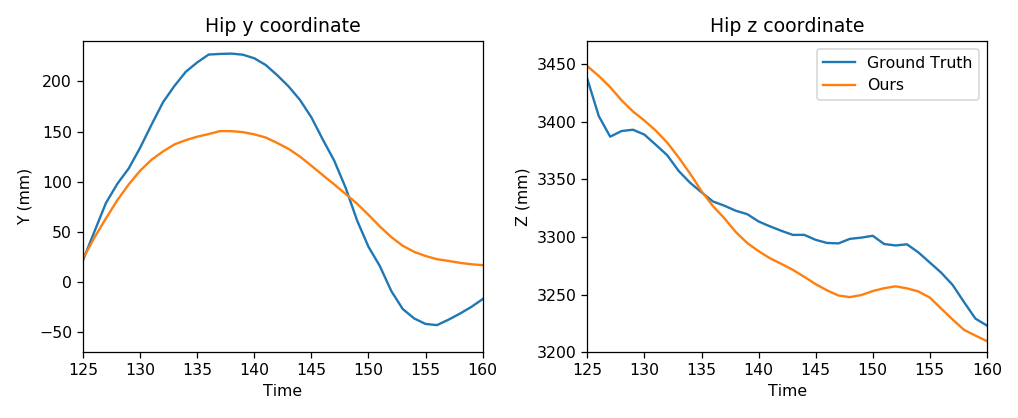}
\end{center}
\caption{\textbf{Trajectory of hip during a jump.} During a jump, the trajectory follows the ground-truth, showing that the model did not overfit to vertical location. Left plot shows the vertical coordinate of the hip in a camera centered coordinate system. The right plot shows the depth of the joint.}
\label{fig:jump}
\end{figure*}
One may assume that the TPN overfitted and simply calculates the depth based on the $y$ coordinates. However, this is mitigated by the fact that the training and test set have different camera viewpoints. Moreover, we applied an augmentation that effectively zooms the the cameras, further increasing the diversity of the inputs (see Section~\ref{sec:imp-details}). We also show an example sequence in Figure~\ref{fig:jump}. It contains the trajectory of the hip of a person jumping backward, taken from MuPoTS Sequence~15. The figure demonstrates, that even though the $y$ coordinate of the joint is increasing, the $z$ coordinate still follows the ground truth. If the TPN was overfitted to the $y$ location, we would expect a jump in the trajectory of the $z$ coordinate

\section{Conclusion}
We proposed a pose refinement approach, that corrects predictions of heavily occluded poses. Our method improves both localization and pose estimation performance, achieving state-of-the-art results on the MuPoTS dataset. We demonstrated that it does not impair performance on unoccluded poses. Our algorithm could be further extended by making the refinement process part of the pose estimation network, in an end-to-end fashion. Also, one drawback of our approach is that it does not include tracking, the combination with a tracking algorithm remains future work.

\section*{Acknowledgment}
 MV received support from the European Union and co-financed by the European Social Fund (EFOP-3.6.3-16-2017-00002). AL was supported by the National Research, Development and Innovation Fund of Hungary via the Thematic Excellence Programme funding scheme under Project no. ED\_18-1-2019-0030 titled Application-specific highly reliable IT solutions. 


\end{document}